\documentclass{article}

    \PassOptionsToPackage{numbers, compress}{natbib}

    \usepackage[preprint]{neurips_2025}

\usepackage{multirow}
\usepackage{colortbl}
\usepackage[utf8]{inputenc} 
\usepackage[T1]{fontenc}    
\usepackage{hyperref}       
\usepackage{url}            
\usepackage{booktabs}       
\usepackage{amsfonts}       
\usepackage{nicefrac}       
\usepackage{microtype}      
\usepackage{xcolor}         
\usepackage{graphicx}

\title{OCSU: Optical Chemical Structure Understanding for Molecule-centric Scientific Discovery}

\author{%
  Siqi Fan\textsuperscript{1} \,
Yuguang Xie\textsuperscript{1} \,
Bowen Cai\textsuperscript{2} \,
Ailin Xie\textsuperscript{2} \,
Gaochao Liu\textsuperscript{1} \\
\bf Mu Qiao\textsuperscript{2} \,
Jie Xing\textsuperscript{2} \,
Zaiqing Nie\textsuperscript{1,2} \thanks{Corresponding author. For any questions or discussions, please email \{fansiqi, dair\}@air.tsinghua.edu.cn.}\\
\textsuperscript{1} Institute for AI Industry Research (AIR), Tsinghua University \quad
\textsuperscript{2} PharMolix Inc.\\
}

\begin{document}

\maketitle

\begin{abstract}
  Understanding the chemical structure from a graphical representation of a molecule is a challenging image caption task that would greatly benefit molecule-centric scientific discovery. Variations in molecular images and caption subtasks pose a significant challenge in both image representation learning and task modeling. Yet, existing methods only focus on a specific caption task that translates a molecular image into its graph structure, i.e., OCSR. In this paper, we propose the Optical Chemical Structure Understanding (OCSU) task, which extends low-level recognition to multilevel understanding and aims to translate chemical structure diagrams into readable strings for both machine and chemist. To facilitate the development of OCSU technology, we explore both OCSR-based and OCSR-free paradigms. We propose DoubleCheck to enhance OCSR performance via attentive feature enhancement for local ambiguous atoms. It can be cascaded with existing SMILES-based molecule understanding methods to achieve OCSU. Meanwhile, Mol-VL is a vision-language model end-to-end optimized for OCSU. We also construct Vis-CheBI20, the first large-scale OCSU dataset. Through comprehensive experiments, we demonstrate the proposed approaches excel at providing chemist-readable caption for chemical structure diagrams, which provide solid baselines for further research. Our code, model, and data are open-sourced at \url{https://github.com/PharMolix/OCSU}.
\end{abstract}

\section{Introduction}
\label{sec:intro}

Molecules serve as the tokens of the language of chemistry, underlying not only chemistry itself but also various scientific fields that rely on chemical information, such as pharmacy, material science, and molecular biology \cite{molgrapher, decimer}. Existing molecular information is scattered across textbooks, publications, and patents. To convey structural information (the spatial arrangement of atoms), molecules are often depicted as 2D images in these documents. This makes \textbf{Optical Chemical Structure Understanding (OCSU)} a crucial component in molecule-centric scientific discovery.

\begin{figure}[t]
\begin{center}
\includegraphics[width=0.95\linewidth]{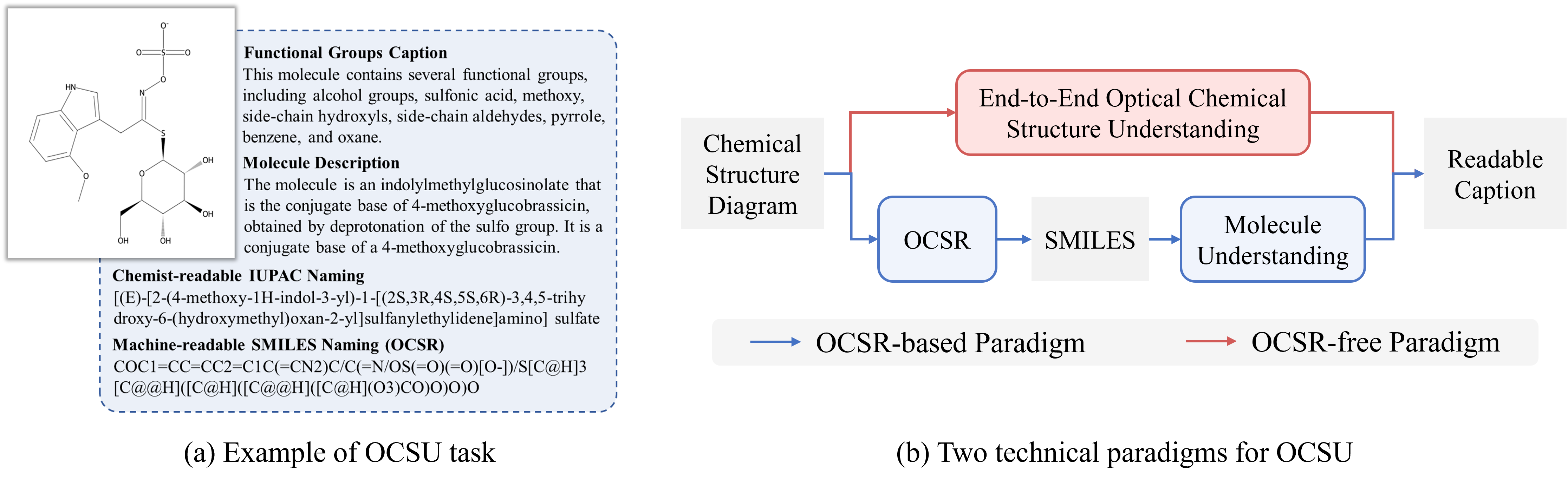}
\end{center}
   \caption{\textbf{Introduction of OCSU.} (a) Example of OCSU task. Optical chemical structure understanding is a special image caption task that describes the molecular diagrams from multiple levels, including four typical subtasks, i.e., functional group caption, molecule description, chemist-readable IUPAC naming, and machine-readable SMILES naming. (b) Two technical paradigms for OCSU. OCSR-based paradigm can fully leverage the power of existing OCSR and SMILES-based task-specific molecule understanding methods, while models are optimized in an end-to-end manner via multitask learning within OCSR-free paradigm.}
\label{fig:ocsu}
\end{figure}

OCSU aims to automatically translate chemical structure diagrams into strings that are readable by chemists or machines, describing the molecule at multiple levels. To formulate the task in a standardized manner, we break down the complex understanding task into three levels: structural motif level, molecule level, and abstract level. This results in four subtasks, namely, functional group caption, molecular description, chemist-readable IUPAC naming, and machine-readable SMILES naming (OCSR), as illustrated in Fig.~\ref{fig:ocsu}. The outputs of OCSU can be readily utilized by both AI-driven drug discovery and large language model technologies to support typical molecule-centric scientific discovery tasks, such as molecule-centric chat, property prediction, and molecule editing.

Technically, the OCSU task faces two main challenges. First, there is the visual representation learning challenge of chemical diagrams, which is caused by variations in drawing styles, image quality, and non-chemical element noise \cite{molscribe, atomLenz}. Second, there are caption task modeling challenges introduced by the various image-text generation subtasks across different caption levels.

To address these challenges, two technical paradigms have emerged, as illustrated in Fig.~\ref{fig:ocsu} (b). One is the two-stage OCSR-based paradigm, and the other is the end-to-end OCSR-free paradigm. In the OCSR-based paradigm, OCSU is decoupled into optical chemical structure recognition (OCSR) and molecule understanding, leveraging the power of existing technologies. Existing OCSR methods \cite{osra, molvec, imago, img2mol, img2graph, molgrapher, decimer, molscribe, atomLenz} have explored various techniques to handle the visual representation learning challenge. Additionally, different task-specific SMILES-based molecule understanding methods \cite{git-mol, momu, molfm, biot5, biot5+} can be utilized for various caption tasks. In contrast, the OCSR-free approach is end-to-end optimized for OCSU and takes advantage of the rapid development of Vision-Language Models (VLMs) \cite{llava, internvl, internvl1.5, internvl2.5, qwenvl, qwen2vl} in robust representation learning and flexible multi-task modeling. However, neither of these paradigms has been systematically discussed and investigated for the OCSU task, which hinders the development of this exciting but challenging research direction.

In this paper, we explore both paradigms and conduct experimental investigations. Our primary aim is to answer 3 key questions regarding OCSU: (1) Does enhancing the performance of submodules lead to improved OCSU performance within the OCSR-based paradigm? (2) Does OCSU benefit from end-to-end multi-task learning within the OCSR-free paradigm? (3) Is one paradigm consistently dominant across all OCSU subtasks?

To provide answers, we first focus on improving OCSR performance. Addressing the issue of false or missing atom detection, we propose \textbf{DoubleCheck}, a method that takes a second look at locally ambiguous atoms and enhances feature attentively. This strengthens atom-level recognition capabilities and improves robustness when dealing with various real-world molecular images. Next, we explore and present \textbf{Mol-VL}, a vision language model, to achieve end-to-end molecular image captioning. Additionally, we construct \textbf{Vis-CheBI20}, the first large-scale OCSU dataset, for model finetuning and systematic evaluation.

Through comprehensive experiments, we demonstrate that both proposed approaches excel at generating chemist-readable captions for chemical structure diagrams and achieve state-of-the-art performance. The DoubleCheck-based approach outperforms other OCSR-based methods, while Mol-VL-7B generally performs better, except in recall performance for the IUPAC naming task. Specifically, in the functional group captioning task, Mol-VL-7B outperforms the strongest baseline approach by 7.7\% in terms of F1 score. We hope our work provides solid baselines and facilitates future development in this field.

Our contributions are summarized as follows:
\begin{itemize}
  \item We propose and formulate the OCSU task, which expands the task scope of molecular structure understanding with chemical diagram input, compared with well-explored OCSR task. We build Vis-CheBI20, the first large-scale OCSU dataset.
  \item We present and explore the two paradigms for OCSU, i.e., OCSR-based and OCSR-free paradigms. We introduce attentive feature enhancement mechanism for local ambiguous atoms to improve OCSR performance in Double-Check and perform end-to-end multi-task learning in Mol-VL.
  \item We validate the effectiveness of the proposed approaches through comprehensive experiments and investigate the performance of the two paradigms on OCSU tasks.
\end{itemize}

\section{Related Work}
\label{sec:related_work}


\subsection{Optical Chemical Structure Recognition (OCSR)}

Optical chemical structure recognition is a well-explored task in the molecular information extraction field, consisting in inferring the structural formulae of a chemical compound based on an image representation of it. Traditional OCSR methods are rule-based, such as MolVec \cite{molvec} and OSRA \cite{osra}. Recently, data-driven deep-learning methods \cite{img2mol, decimer, molscribe, molscribe, atomLenz} have been proposed to handle the visual representation learning challenge of chemical diagrams. These existing methods performing OCSR take the image as input and predict the SMILES (Simplified Molecular Input Line Entry System) representation \cite{smiles} of the molecule. Nevertheless, SMILES representation is machine readable strings, while it is not chemist-friendly. This hampers overall understanding of the optical chemical structure and hinders further application of cutting-edge natural language processing approaches, e.g., LLM, for molecule-centric scientific discovery. Our work expands low-level recognition to multilevel understanding and aims to translate chemical structure diagrams into readable strings for both machine and chemist. 

\subsection{Vision-Language Models (VLMs)}

With blooming development in multimodal large language model, vision-language models (VLMs) have been studied to confront the challenging vision-based cross-modal understanding tasks, including image caption, visual question answering, and visual reasoning. The development of VLMs has greatly improved the performance of visual understanding, and the pre-trained VLMs have been widely employed in various fields. Visual document understanding (VDU) is a typical application for VLMs, which aims to comprehend images with rich text information \cite{pix2struct, ureader, docowl}. A series of VLMs are end-to-end finetuned for VDU task and achieve better performances without relying on an off-the-shelf OCR system \cite{ureader, docowl, docpedia}. VDU for scientific articles are more challenging because of the specialized diagrams like chemical figures. In this work, we first adopt VLMs for OCSR-free end-to-end optical chemical structure understanding and hope to facilitate the future development of the exciting but challenging OCSU task.

\section{Preliminary}

\subsection{Problem Formulation}

Optical Chemical Structure Understanding (OCSU) is technically an image captioning task that automatically extracts structural information from molecular diagrams and translates the input image into chemist-readable or machine-readable strings. Aiming to facilitate molecule-centric scientific discovery, the output strings are either natural language descriptions or the widely used molecule representation, SMILES. These outputs can be further utilized as inputs for downstream tasks, such as molecule question-answering, property prediction, and molecule editing.

Compared to the well-explored OCSR, we expand the scope and make OCSU a more general task. The caption tasks are grouped into three levels: functional group caption for the motif level, general description for the molecule level, and IUPAC/SMILES naming for the abstract level, as described in Table~\ref{tab:definition}. Functional group caption is the most fine-grained understanding task in OCSU, focusing on functional substructures that contribute to the molecule's chemical or biomedical functions. Molecule description provides a general profile of the molecule based on its structure. The two naming tasks either output chemist-readable IUPAC names based on motifs or generate machine-readable SMILES strings based on atoms.

OCSU task is formulated as $P(\mathcal{T}|\mathcal{I})$, where $\mathcal{I}$ and $\mathcal{T}$ indicate image and text respectively. In OCSR-free paradigm, an unified VLM is utilized to model $P(\mathcal{T}|\mathcal{I})$ in an end-to-end manner. Meanwhile, it is separated into two cascaded subtask in OCSR-based paradigm, i.e., optical chemical structure recognition $P(\mathcal{S}|\mathcal{I})$ and SMILES-based molecule understanding $P(\mathcal{T}|\mathcal{S})$, formulated as $P(\mathcal{T}|\mathcal{I}) = P(\mathcal{S}|\mathcal{I}) P(\mathcal{T}|\mathcal{S})$, where $\mathcal{S}$ indicates SMILES.

\begin{table*}
  \centering
  \caption{\textbf{Definition of optical chemical structure understanding (OCSU).}}
  \begin{tabular}{p{1.2cm}p{3cm}p{8.2cm}} 
  \hline
    \textbf{Level}          & \textbf{Task}                 & \textbf{Description} \\ \hline

    \multirow{1}*{Motif}      & FuncGroup Caption      & Describe the composition and list major structural units.\\ \hline
    \multirow{1}*{Molecule}   & Molecule Description          & Describe the molecule based on the structure.\\ \hline
    \multirow{2}*{Abstract}   & IUPAC Naming                  & Translate chemical structure diagram into chemist-readable representation according to motif-level structure.\\
        \cline{2-3}                 & SMILES Naming                 & Translate chemical structure diagram into machine-readable representation according to atom-level structure.\\ \hline    
    
  \end{tabular}
  \label{tab:definition}
\end{table*}

\subsection{Vis-CheBI20 Dataset}
\label{sec:dataset}

\begin{table*}
    \centering
    \caption{\textbf{Vis-CheBI20 dataset statistics.}}
    \begin{tabular}{c|cc|cc}
    \hline
    \textbf{Input}          & \multicolumn{2}{c|}{\textbf{Tasks}} & \textbf{TrainingSet}    & \textbf{TestSet} \\ \hline
    \multirow{4}*{Images}   & \multirow{2}*{Motif Level}   & Functional Group Recognition  & 26,144 & 3,269 \\ \cline{3-5}
                            &                              & Functional Group Caption      & 26,144 & 3,269 \\ \cline{2-5}
                            & Molecule Level               & Molecule Description          & 26,407 & 3,300 \\ \cline{2-5}
                            &\multirow{2}*{Abstract Level} & IUPAC Naming                  & 26,200 & 2,680 \\ \cline{3-5}
                            &                              & SMILES Naming                 & 26,407 & 3,300 \\ \hline    
      
    \end{tabular}
    \label{tab:vis-size}
\end{table*}

We build Vis-CheBI20, a large-scale dataset comprising 29.7K molecular diagrams and 117.7K OCSU image-text pairs, to facilitate fine-tuning and evaluation. We provide an overview of our dataset in Tab.~\ref{tab:vis-size}. The construction process involves the following steps:

\textbf{Raw data collection.} The primary source of Vis-CheBI20 is ChEBI-20 \cite{text2mol}, a widely adopted molecule dataset for text-molecule translation. We directly utilize molecule description pairs to ensure that the OCSU benchmark at the molecule level on Vis-CheBI20 is comparable to existing molecule understanding methods on ChEBI-20 that use SMILES as input. To collect other essential data, including functional groups and IUPAC names, we employ professional tools and databases. Specifically, we retrieve IUPAC names from the PubChem database \cite{PubChem} using the corresponding SMILES as queries. We also define an expert-reviewed scope of functional groups relevant to practical molecule-centric discovery, such as drug discovery. This scope includes 59 general and 106 cyclic functional groups. The functional groups for each molecule are automatically extracted using RDKit \cite{rdkit}.

\textbf{Image-Text pairs generation.} We follow the general molecular diagram generation process used in real-world scenarios, such as in journals and patents. All images are generated using RDKit with SMILES as input. This step ensures that performance on Vis-CheBI20 reflects capabilities in practical applications. We design question–answer template for each OCSU task and use the collected raw data to generate image–text pairs. In addition to the four typical tasks, we introduce functional group recognition as an auxiliary task in Vis-CheBI20, which is discussed in Sec.~\ref{sec:molvl}. Examples of Vis-CheBI20 are provided in the Appendix~\ref{sec:appendix_dataset}.

\textbf{Evaluation metircs.} We adopt different evaluation metrics for each OCSU task. (1) Functional group caption is an information retrieval task, we use the F1 score as the main metric to evaluate the retrieval performance of functional groups. (2) Molecule description is a natural language generation task, we adopt the evaluation metrics used in ChEBI-20, including BLEU-2, BLEU-4, ROUGE-2, ROUGE-L, and METEOR. Additionally, we also use BERTScore-F1 to measure the semantic similarity. (3) IUPAC naming is also a natural language generation task, we use general metrics, such as BLEU-2, BLEU-4, ROUGE-2, ROUGE-L, and METEOR, to measure the similarity. (4) SMILES naming is a well-explored task, we adopt the exact matching accuracy, which is widely used in OCSR benchmark.

\section{Method}
\label{sec:method}

In this section, we explore both OCSR-based and OCSR-free paradigms. Specifically, (1) we introduce attentive feature enhancement mechanism for local ambiguous atoms to enhance the OCSR performance in DoubleCheck and (2) perform end-to-end multi-task learning in Mol-VL.

\subsection{DoubleCheck for OCSR-based Paradigm}

\begin{figure}[t]
    \centering
    \includegraphics[width=0.99\linewidth]{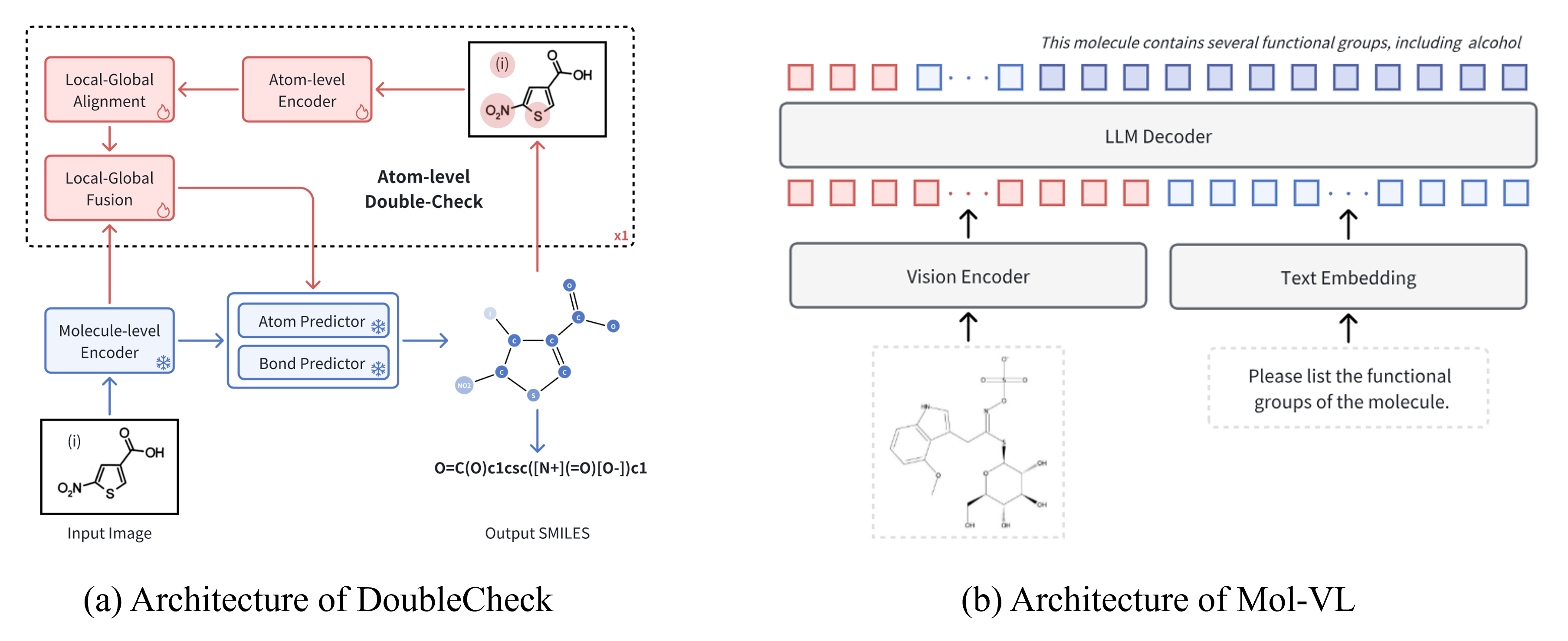} 
    \caption{\textbf{Exploration on OCSR-based and OCSR-free paradigms for OCSU.} (a) Architecture of DoubleCheck. An attentive feature enhancement module is introduced for local ambiguous atoms. (b) Architecture of Mol-VL. A vision-language model is end-to-end optimized via multi-task learning.}
    \label{fig:method}
\end{figure}

Following \cite{molscribe, img2graph}, we handle OCSR in a molecular graph reconstruction way. The image-to-graph translation is formulated as a conditional generation process: 

\begin{equation}
     P(\mathcal{S}|\mathcal{I}) = P(\mathcal{A}|\mathcal{I}) P(\mathcal{B}|\mathcal{A},\mathcal{I})
\end{equation}
where $\mathcal{A}$ and $\mathcal{B}$ indicate atom and bond, while $P(\mathcal{A}|\mathcal{I})$ and $P(\mathcal{B}|\mathcal{A},\mathcal{I})$ are parametrized as an atom predictor and a bond predictor. The atom predictor is an autoregressive decoder that generate atoms in a sequence, while the bond predictor is a feedforward network that predicts the bond between each pair of atoms \cite{molscribe}.

\begin{equation}
     P(\mathcal{A}|\mathcal{I}) = \prod_{i=1}^{n} P(a_i|\mathcal{A}_{<i},\mathcal{I})
\end{equation}

\begin{equation}
     P(\mathcal{B}|\mathcal{A}, \mathcal{I}) = \prod_{i=1}^{n} \prod_{j=1}^{n} P(b_{i,j}|\mathcal{A},\mathcal{I})
\end{equation}
Since the chemical bond type is limited (usually include seven types), accurate bond prediction is easier to achieve. While suffering from various drawing style and non-atom noise, atom prediction is more challenging and limit the molecular graph reconstruction performance. We manage to enhance the recognition capability via local ambiguous atom double check and attentive local-global feature enhancement, as shown in Fig.~\ref{fig:method}.


\textbf{Local Ambiguous Atom Double Check} is to answer where to enhance. We determine the ambiguous atom according to the atom prediction confidence. A 2D gaussian mask is positioned at the center of the ambiguous atom and identifies a region of interest for double check. We adopt Swin-B \cite{swin} as the atom-level encoder $\Phi_{l}$, which theoretically only encodes the selected region, as the rest of the image is not visible after image masking. 

\begin{equation}
    \mathcal{F}_{l} = \mathbf{\Phi_{l}} (\mathcal{I}_{aa})
\end{equation}
where $\mathcal{I}_{aa}$ is the masked image, and $\mathcal{F}_{l}$ is the encoded local feature.

\textbf{Attentive Local-Global Feature Enhancement} is to answer how to enhance. The learned local feature $\mathcal{F}_{l}$ is fused with molecular global feature $\mathcal{F}_{g}$ for feature enhancement. Since molecule-level encoder $\Phi_{g}$ and atom-level encoder $\Phi_{l}$ are trained individually, a feature alignment step is introduced before fusion. We use a two-layer MLP for feature alignment, and another two-layer MLP is employed for weight generation. The original molecule-level global feature $\mathcal{F}_{g}$ is enhanced with the aligned atom-level local feature $\hat{\mathcal{F}_{l}}$ via weighted summation.

\begin{equation}
    \hat{\mathcal{F}_{l}}= \mathbf{MLP}(\mathcal{F}_{l})
\end{equation}
\begin{equation}
    \mathcal{F}_{e} = \mathcal{F}_{g} + \mathbf{MLP}(\mathcal{F}_{g} \oplus \hat{\mathcal{F}_{l}}) * \hat{\mathcal{F}_{l}}
\end{equation}

\textbf{Training strategy for DoubleCheck.} We design a two-stage training strategy. We train molecule-level encoder, atom predictor, and bond predictor in the first stage. In the second stage, we focus on ambiguous atom recognition and train atom-level encoder, local-global alignment module, and local-global fusion module. To manually introduce ambiguous atom areas for training, we randomly add 2D gaussian mask noise and generate the corresponding $\mathcal{I}_{aa}$ to obtain the image pairs.

\subsection{Mol-VL for OCSR-free Paradigm}
\label{sec:molvl}

Mol-VL is proposed to model $P(\mathcal{T}|\mathcal{I})$ in an end-to-end manner.

\textbf{Model Architecture.} As illustrated in Fig.~\ref{fig:method}, Mol-VL is composed of a vision encoder and a large language model. Following \cite{qwen2vl, internvl}, we adopt Vision Transformer (ViT) as the encoder, and both naive dynamic resolution \cite{NaiveDynamicResolution} and multimodal rotary position embedding (M-RoPE) \cite{qwen2vl} are utilized. The vision encoder takes molecular diagrams $\mathcal{I}$ as inputs and generates feature embeddings. The LLM is a decoder-based transformer model and outputs natural language $\mathcal{T}$. 

\textbf{Multi-task learning strategy.} To take advantage of existing pretrained VLMs, we adopt the pretrained weights and further finetuned it on Vis-CheBI20 training set via full-parameter supervised fine-tuning. To alleviate the catastrophic forgetting problem, we organize the virtual token embeddings for molecular diagrams and the question in an instruction-following manner. The system prompt \textit{`You are working as an excellent assistant in chemistry and molecule discovery. Below a human gives the representation of a molecule. Answer a question about it.'} identify the role of LLM as a molecule-centric research assistant, and special tokens $<image>$ and $</image>$ are introduced to help LLM understand where the sturctural feature of molecular diagram start and end. The prompt allows the LLM answer OCSU questions by reasoning over molecular diagrams in an auto-agressive manner and maintain the original capability given different instructions. In addition to the 4 typical tasks in OCSU, Mol-VL is also optimized on the functional group recognition as an auxiliary task. The model is asked to recognize the highlighted functional group to let it better understand the structural context from molecular diagrams. Example of the auxiliary task is provided in the Appendix~\ref{sec:appendix_dataset}.

\section{Experiments}
\label{sec:experiments}

In this section, we evaluate the performance of the two paradigms on Vis-CheBI20 and demonstrate that the proposed approaches excel at providing chemist-readable caption for chemical structure diagrams. We start with a brief introduction of our training setups, followed by detailed evaluations on core OCSU tasks. We also present an in-depth analysis of the experimental results.

\subsection{Training Setup}
\label{sec:implementation}
For DoubleCheck, we employ the architecture of MolScribe \cite{molscribe} as backbone and train the model with the same training data used in \cite{molscribe} for $30$ epochs in the first stage. We further optimize the model for additional $10$ epochs using the augmented training data with the randomly introduced 2D gaussian noise. We use a maximum learning rate of 4e-4 with a linear warmup for $5\%$ steps and a cosine function decay. The training process takes 4 days on 4 NVIDIA A100 GPUs.
For Mol-VL, we adopt Qwen2-VL (2B and 7B version)  \cite{qwen2vl} as the base model and further finetuned for $50$ epochs with a maximum learning rate of 1e-5. We perform linear warmup for the first $10\%$ steps and a cosine annealing strategy to stabilize training. The training process takes 10 days on 4 A100 GPUs.

\subsection{Performance Evaluation on Molecule Description}

\begin{table*}[t]
\centering
\caption{\textbf{Performance on Molecule Description.} BL: BLEU. R: ROUGE. BS: BERTScore.}
\begin{tabular}{c|c|cccccc}
\hline
Methods                                 & Input & BL-2     & BL-4     & R-2       & R-L       & METEOR   & BS-F1\\ \hline
\multicolumn{8}{c}{Molecule Input (Single-task Specialist Models)} \\ \hline
MoMu \cite{momu}                        & SMILES & 54.9    & 46.2     & 47.9      & 57.5      & 57.6     & -\\
MolFM \cite{molfm}                      & SMILES & 58.5    & 49.8     & 50.8      & 59.4      & 60.7     & -\\
BioT5 \cite{biot5}                      & SMILES & 63.5    & 55.6     & 55.9      & 63.3      & 65.6     & - \\
MolCA \cite{molca}                      & SMILES & 62.0    & 53.1     & 53.7      & 61.8      & 65.1     & - \\
BioT5+ \cite{biot5+}                    & SMILES & \bf66.6 & \bf59.1  & \bf58.4   & \bf65.0   & \bf68.1  & - \\ \hline
\multicolumn{8}{c}{Molecule Input (LLM-based Generalist Models)} \\ \hline
3D-MoLM \cite{3dmollm}                  & SMILES & 6.7     & 3.0      & 4.2       & 8.6       & 18.3     & - \\
Mol-Instructions \cite{molinstructions} & SMILES & 24.9    & 17.1     & 20.3      & 28.9      & 27.1     & - \\
BioMedGPT \cite{biomedgpt}              & SMILES & 30.6    & 19.8     & 25.7      & 38.3      & 35.0     & - \\ 
GIT-Mol \cite{git-mol}                  & SMILES & \bf35.2 & \bf26.3  & \bf48.5   & \bf56.0   & \bf43.0  & - \\ \hline
\multicolumn{8}{c}{Vision Input} \\ \hline
MolScribe \& BioT5+                     & Image & 53.43    & 44.64    & 46.48     & 55.30     & 57.09    & 59.67 \\
\textbf{DoubleCheck \& BioT5+}         & Image & 54.40    & 45.56    & 47.21     & 55.94     & 58.07    & 60.62 \\
Qwen2-VL-7B \cite{qwen2vl}              & Image & 6.34     & 1.13     & 4.53      & 14.15     & 19.95    & 10.82 \\
\textbf{Mol-VL-2B}                      & Image & 50.57    & 40.26    & 41.99     & 52.25     & 53.34    & 56.50 \\
\textbf{Mol-VL-7B}                      & Image & \bf55.73 & \bf46.14 & \bf47.26  & \bf56.61  & \bf58.14 & \bf61.51 \\ \hline
\end{tabular}
\label{tab:performance_mol_desp}
\end{table*}

Molecule description generation is a classic molecule understanding task, and several previous works have explored SMILES-based description generation \cite{momu, molfm, biot5, molca, biot5+}. With the rise of LLM-based applications in various vertical tasks, LLM-based approaches have also been employed for molecule description with SMILES input \cite{3dmollm, molinstructions, biomedgpt, git-mol}. We compare our methods with both task-specific and LLM-based generalist models in Tab.~\ref{tab:performance_mol_desp}. Generally, task-specific models outperform LLM-based generalist models on the molecule description task. For the OCSU setting, our proposed methods achieve comparable performance with MoMu \cite{momu}, significantly outperforming Qwen2-VL-7B \cite{qwen2vl}.

Considering the differences between vision input and direct molecule input, we further adopt Molscribe \cite{molscribe} and BioT5+ \cite{biot5+} as baselines, both of which are state-of-the-art methods in their respective areas. With better OCSR performance (Sec.~\ref{sec:in_depth}), our DoubleCheck method, in collaboration with BioT5+, surpasses the baseline by an average of $0.81\%$. Mol-VL-7B achieves state-of-the-art performance on vision-based molecule description, while the 2B variant underperforms the OCSR-based baseline. The experimental results demonstrate that general VLMs cannot directly achieve promising performance on the molecule diagram understanding task, but their performance can be significantly boosted through further multi-task fine-tuning. We speculate that scaling the training dataset will bring substantial improvements, which we reserve for future exploration.

\subsection{Performance Evaluation on Functional Group Caption}

We adopt MolScribe in conjunction with RDKit \cite{rdkit} as the baseline for the OCSR-based paradigm. When employing SMILES-based chemical tools, the functional group captioning task is converted into a substructure matching problem. Qwen2-VL is assessed as the baseline for the OCSR-free paradigm, yet it struggles significantly with the captioning task, achieving an extremely low F1 score of 7e-5\%. In addition to the F1 score, we also present the precision and recall metrics in Tab.~\ref{tab:performance_func_cap}.

The enhanced molecule recognition capability yields a notable improvement in performance. The combination of DoubleCheck and RDKit surpasses the baseline by an absolute gain of $4.03\%$. We observe that end-to-end optimization effectively enables the OCSR-free approach to attain superior performance. Mol-VL-7B demonstrates a performance advantage of $3.69\%$ over the OCSR-based approach. As a substructure-level pattern within a molecule, the functional group (structural context) can be better learned through end-to-end multi-task learning. In contrast, OCSR methods, which focus more on atom-level patterns than on structural context, may not achieve optimal functional group caption. We present error analysis of Mol-VL-7B in Appendix~\ref{sec:error}. 

\subsection{Performance Evaluation on IUPAC Naming}

\begin{table}[t]
\centering
\caption{\textbf{Performance on Functional Group Caption.}}
\begin{tabular}{c|c|cc}
\hline
Model & F1 & Precision & Recall \\ \hline
MolScribe \& RDKit & 89.60 & 91.88 & 87.87 \\
\textbf{DoubleCheck \& RDKit} & 93.63 & 93.58 & 93.90 \\
\textbf{Mol-VL-2B} & 95.40 & 95.48 & 95.99 \\
\textbf{Mol-VL-7B} & \textbf{97.32} & \textbf{96.94} & \textbf{98.15} \\ \hline
\end{tabular}
\label{tab:performance_func_cap}
\end{table}

\begin{table*}[h]
\centering
\caption{\textbf{Performance on IUPAC Naming.} BL: BLEU. R: ROUGE.}
\begin{tabular}{c|cccccc}
\hline
Model & BL-2 & BL-4 & R-1 & R-2 & R-L & METEOR \\ \hline
MolScribe \& PubChemDB & 61.08 & 61.03 & 82.73 & 81.61 & 82.68 & 82.70 \\
\textbf{DoubleCheck \& PubChemDB} & 64.28 & 64.23 & 84.45 & \bf83.26 & \bf84.40 & \bf84.47 \\
Qwen2-VL \cite{qwen2vl} & 14.87 & 6.27 & 31.44 & 9.10 & 28.48 & 29.14 \\
\textbf{Mol-VL-2B} & 80.94 & 72.79 & 77.67 & 60.21 & 72.81 & 77.76 \\
\textbf{Mol-VL-7B} & \bf83.72 & \bf76.95 & \bf81.88 & 66.81 & 77.69 & 81.72 \\ \hline
\end{tabular}
\label{tab:performance_iupac}
\end{table*}

We also employ MolScribe-based approach and Qwen2-VL as baselines for the IUPAC naming task. Following the strategy in dataset construction, we query the PubChem database using the SMILES generated by MolScribe to obtain the corresponding IUPAC names. The performance comparison is reported in Tab.~\ref{tab:performance_iupac}.
As expected, the DoubleCheck-based approach achieves superior performance, with an average absolute gain of $2.21\%$. The explicit IUPAC retrieval relies heavily on accurate SMILES queries, and our method outperforms MolScribe in SMILES generation (as discussed in Sec.~\ref{sec:in_depth}). It is noteworthy that OCSR-based and OCSR-free approaches each have performance advantages in different aspects—precision and recall, respectively. The Mol-VL series demonstrate state-of-the-art performance in generation precision, as measured by BLEU metrics, while they lag behind the OCSR-based approach in recall, as indicated by ROUGE metrics. We hypothesize that a systematic ensemble approach could bring overall performance improvements in this task.

\subsection{In-depth Analysis}
\label{sec:in_depth}
\begin{figure}[h]
    \centering
    \includegraphics[width=0.98\linewidth]{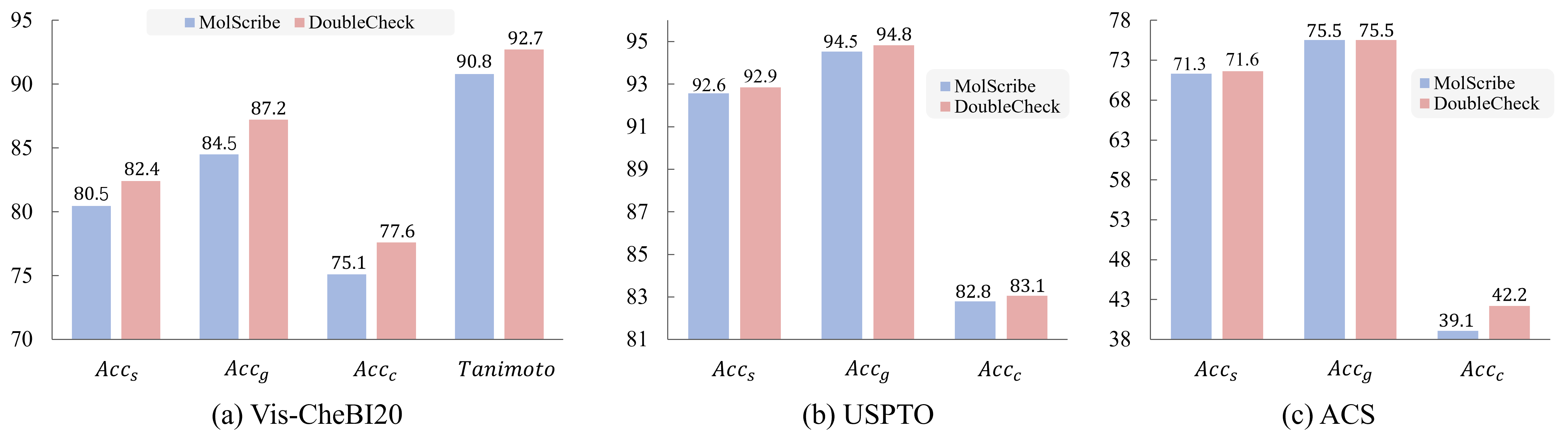} 
    \caption{\textbf{Performance evaluation on OCSR.} (1) Performance on Vis-CheBI20. The performance advantage of DoubleCheck demonstrate the effectiveness of the proposed feature enhancement mechanism. (2) Performance on USPTO. DoubleCheck outperforms MolScribe on real-world patent scenario. (3) Performance on ACS. DoubleCheck surpasses MolScribe on real-world journal scenario.}
    \label{fig:ocsr}
\end{figure}

\textbf{Abation study on DoubleCheck.} We propose DoubleCheck to improve OCSR performance by attentively enhancing the features of locally ambiguous atoms. Since DoubleCheck employ the architecture of MolScribe as backbone, we further compare the OCSR performance of them for ablation study. We report four evaluation metrics in Fig.~\ref{fig:ocsr}(a): exact matching accuracy ($Acc_s$), accuracy without considering chirality ($Acc_g$), accuracy on chiral molecules ($Acc_c$), and Tanimoto similarity. DoubleCheck outperforms MolScribe across all metrics, with an average improvement of $2.27\%$ ($+1.97\% Acc_s$, $+2.70\% Acc_g$, $+2.49\% Acc_c$, and $+1.91\% Tanimoto$). In addition, DoubleCheck also shows performance advantage on a perturbed patent dataset (Appendix~\ref{sec:appendix_ocsr}). The experimental results demonstrates the effectiveness of the proposed attentive feature enhancement mechanism.

\textbf{Further study on Vis-CheBI20.} Although we follow the general diagram generation process in real-world scenario, we are wondering whether the performance on Vis-CheBI20 is consistent with that on other realistic dataset. Therefore, we further evaluate DoubleCheck on several widely adopted benchmarks with real patent and journal molecular images, including \textit{USPTO} \cite{uspto} and \textit{ACS} \cite{molscribe}. We report the three accuracy metrics in Fig.~\ref{fig:ocsr}(b-c). Consistent with that on Vis-CheBI20, DoubleCheck performs better than MolScribe, and achieves an absolute gain of $3.12\%$ on chiral molecules of ACS publications. The experimental results demonstrate the effectiveness of our proposed Vis-CheBI20, which can reflect the performance on real-world scenarios.

\begin{figure}[h]
    \centering
    \includegraphics[width=0.98\linewidth]{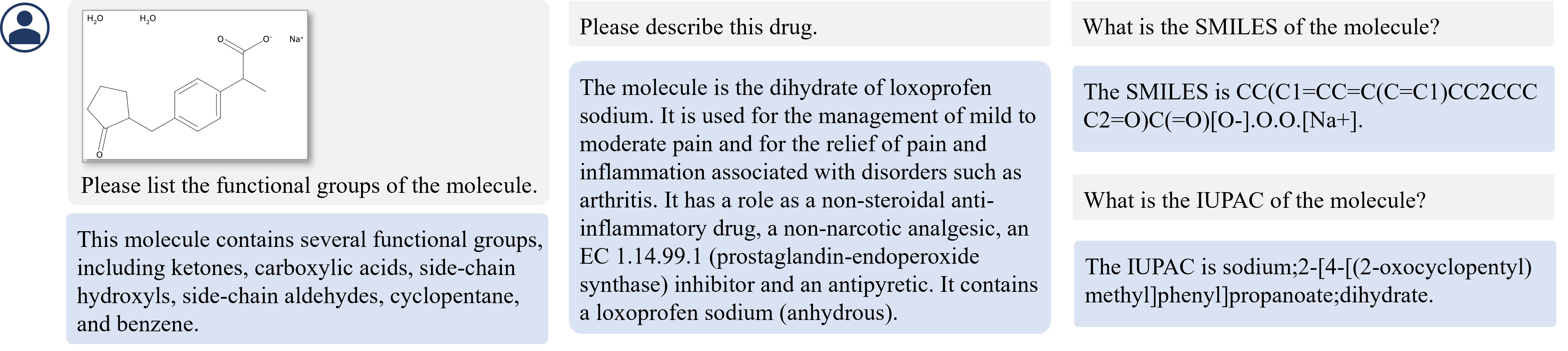} 
    \caption{\textbf{A qualitative example of Mol-VL-7B and an application example in practical scenario.}}
    \label{fig:qualitative_example}
\end{figure}

\textbf{Qualitative analysis of Mol-VL.} We provide a qualitative example of Mol-VL-7B and present a practical scenario of OCSU, as shown in Fig.~\ref{fig:qualitative_example}. A user uploads a chemical diagram and expresses curiosity about the substructures within the molecule. The assistant responds by listing the functional groups present. This prompts the user to inquire about general descriptions of the molecule. The assistant then provides detailed information on its usages and roles. Subsequently, the user can utilize the generated SMILES and IUPAC names for further molecule-centric scientific discovery.

\section{Discussion}
\label{sec:discussion}

We explore and evaluate approaches in both OCSR-based and OCSR-free paradigms. Benefiting from the widespread use of SMILES representation, the OCSR-based approach can leverage the power of existing SMILES-based molecule understanding methods with predicted SMILES, directly supporting downstream tasks such as property prediction and molecule editing. \textit{Enhancing the OCSR submodule can lead to improved OCSU performance.} Meanwhile, OCSR-free approach can support multiple OCSU tasks with a single unified model. \textit{OCSU can benefit from end-to-end multi-task learning.}  Optimized with multi-level OCSU tasks, the model is guided to model the molecular structure from local motifs to the global molecule and then to abstract naming, which is beneficial for structure-based reasoning. Additionally, the model is encouraged to learn more intrinsic and robust patterns to achieve a deeper understanding of the chemical structure. \textit{Technically, both paradigms have their pros and cons, and we present an experimental analysis in Sec.~\ref{sec:experiments}.} Theoretically, the OCSR-free approach has greater potential. The multimodal input of VLMs can better address the more challenging OCSU tasks, such as Markush understanding and reasoning, which needs further research in this field. We are going to enrich Vis-CheBI20 with more diverse tasks and explore more challenging tasks in future work. We declare that OCSU should be restricted to research purposes, and any further applications in molecule discovery should undergo comprehensive experiments.

\section{Conclusion}

In this work, we introduce a novel image captioning task, Optical Chemical Structure Understanding (OCSU), aiming at advancing molecule-centric scientific discovery. OCSU extends the scope of the task from low-level recognition to multi-level understanding, enabling the generation of both machine-readable and chemist-readable strings. We formally define the problem and construct the first large-scale dataset, Vis-CheBI20, for fine-tuning and evaluation purposes. We explore both technical paradigms — OCSR-based and OCSR-free — resulting in the development of DoubleCheck for enhanced OCSR and Mol-VL for end-to-end OCSU. We provide comprehensive experimental results to evaluate the proposed approaches and demonstrate their effectiveness. We anticipate that the exploration of OCSU paradigms and the public release of our dataset will pave the way for novel research avenues and practical applications in this field.

{
    \small
    \bibliographystyle{ieeenat_fullname}
    \bibliography{main}
}




\appendix

\section{Details of Vis-CheBI20}
\label{sec:appendix_dataset}

\begin{table*}
    \centering
    \caption{\textbf{Data examples of Vis-CheBI20 for OCSU.}}
    \begin{tabular}{c|m{10cm}}
    \hline
    Input Image & Question-Answer Pairs \\ \hline
    \multirow{29}*{\includegraphics[width=0.2\linewidth]{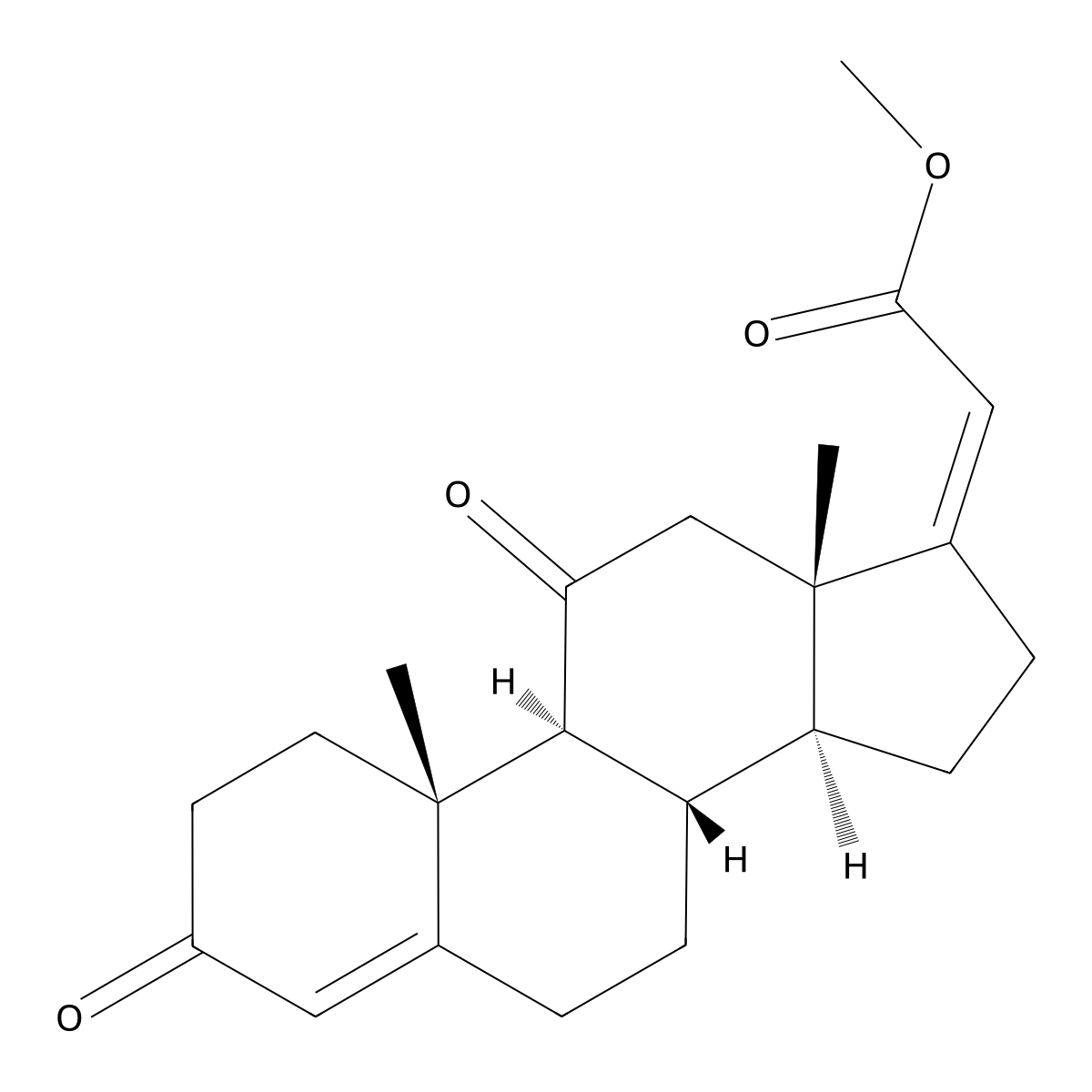}} &  \\
    & \textbf{Functional Groups Caption} \\ 
    &  \\
    & [Question] Please list the functional groups of the molecule. \\
    & [Answer] This molecule contains several functional groups, including esters, ketones, carbonyl methyl ester, methoxy, side-chain aldehydes, alkenes, cyclopentane, cyclohexane, cyclohexene, and 1,2,6,7,8,9,10,11,12,13,14,15,16,17-tetradecahydrocyclopenta[a] phenanthren-3-one. \\
    &  \\ \cline{2-2}
    &  \\
    & \textbf{Molecule Description} \\ 
    &  \\
    & [Question] Please describe this drug. \\
    & [Answer] The molecule is a steroid ester that is methyl (17E)-pregna-4,17-dien-21-oate substituted by oxo groups at positions 3 and 11. It is a 3-oxo-Delta(4) steroid, an 11-oxo steroid, a steroid ester and a methyl ester. It derives from a hydride of a pregnane. \\ 
    &  \\ \cline{2-2}
    &  \\
    & \textbf{IUPAC Naming} \\ 
    &  \\
    & [Question] What is the IUPAC of the molecule? \\
    & [Answer] The IUPAC is methyl(2Z)-2-[(8S,9S,10R,13S,14S)-10,13-dimethyl-3,11-dioxo-1,2,6,7,8,9,12,14,15,16-ecahydrocyclopenta[a]phenanthren-17-ylidene]acetate. \\ 
    &  \\ \cline{2-2}
    &  \\
    & \textbf{SMILES Naming} \\ 
    &  \\
    & [Question] What is the SMILES of the molecule? \\
    & [Answer] The SMILES is C[C@]12CCC(=O)C=C1CC[C@@H]3[C@@ H]2C(=O)C[C@]$\backslash$4([C@H]3CC/C4=C/C(=O)OC)C. \\
    &  \\ \hline
    \multirow{4}*{\includegraphics[width=0.2\linewidth]{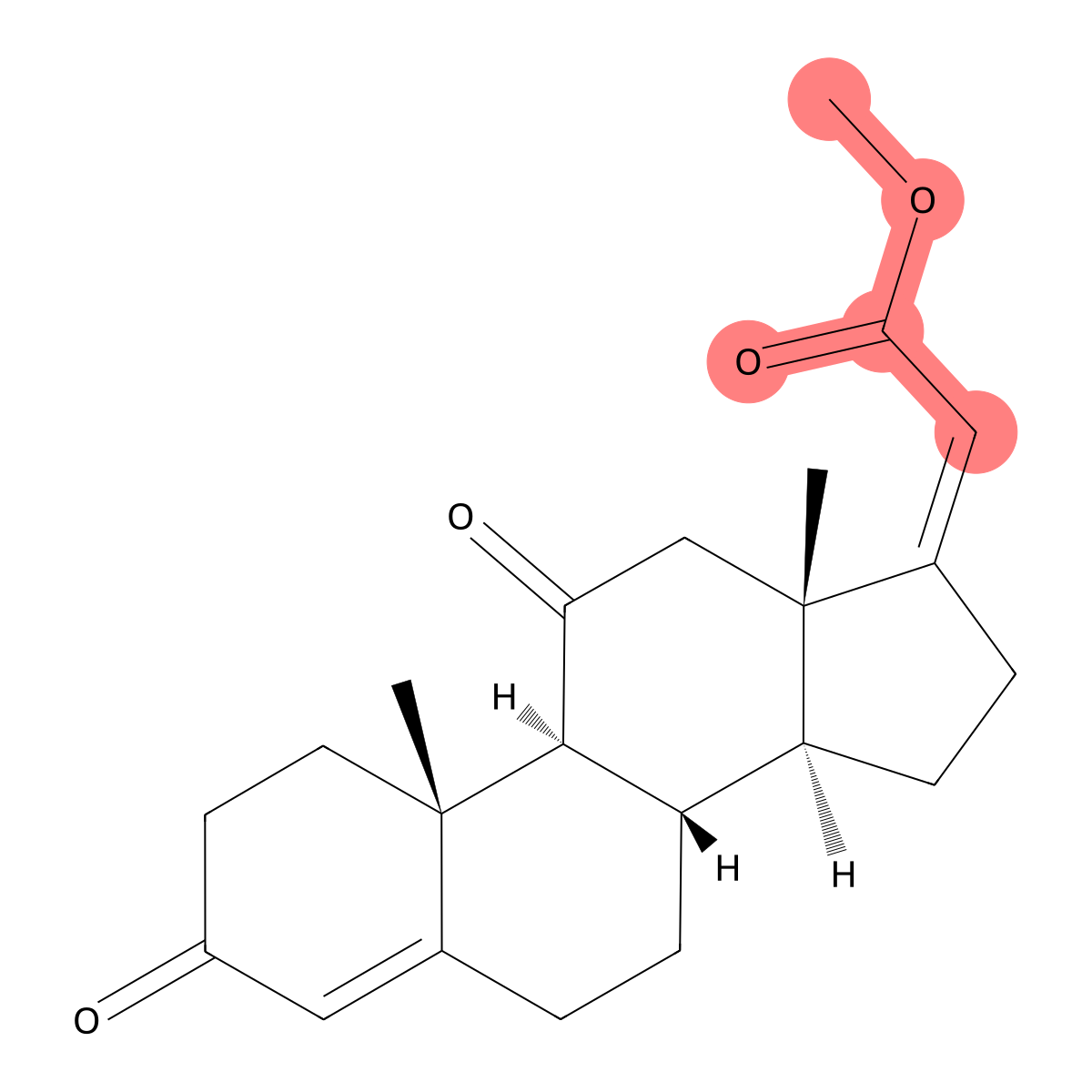}} &  \\
    &  \\
    & \textbf{Functional Groups Recognition} \\ 
    &  \\
    & [Question] Which functional group is highlighted? \\
    & [Answer] The functional group esters is highlighted. \\ 
    &  \\ 
    &  \\ \hline
    \end{tabular}
    \label{tab:dataset_example}
\end{table*}

\subsection{Data Examples}

We present data examples from the Vis-CheBI20 dataset, as shown in Tab.~\ref{tab:dataset_example}.

For each molecule diagram, we generate image-text pairs corresponding to four typical OCSU tasks: functional group captioning, molecule description, IUPAC naming, and SMILES naming. We develop question-answer templates for each task and use the collected raw data to create these pairs. Additionally, functional group recognition is implemented as an auxiliary task to aid model fine-tuning. The model is tasked with identifying the highlighted functional groups, which helps it better understand the structural context within molecular diagrams.

\subsection{Statistical Analysis}

\begin{figure}[t]
    \centering
    \includegraphics[width=0.99\linewidth]{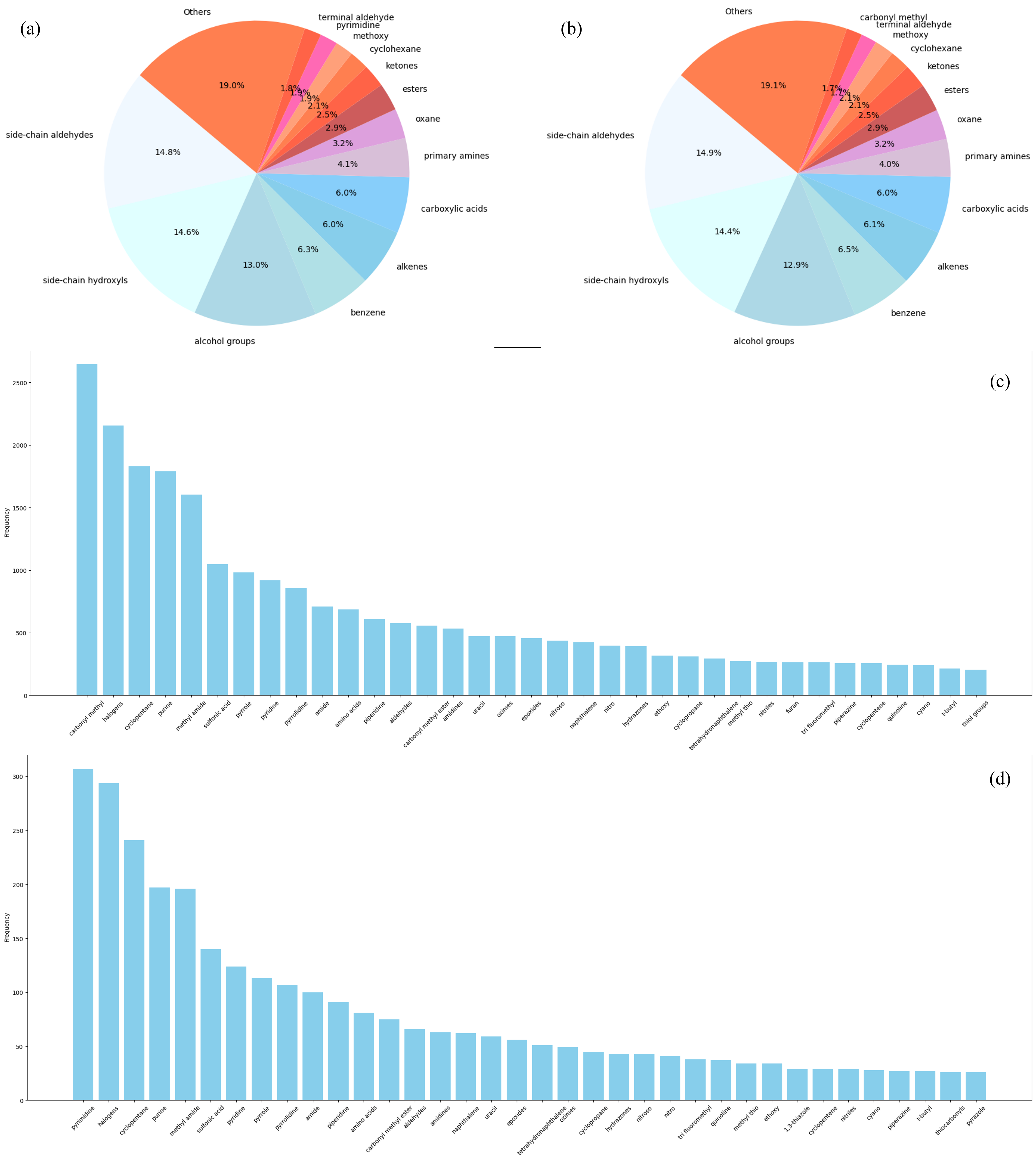} 
    \caption{\textbf{Statistical analysis on functional groups.} We have separately counted the distribution of functional groups in the training set and the test set. The most common functional groups in the training set and the test set are shown in (a) and (b), respectively. For the “others” category, we further counted and displayed the top 35 most frequent ones in (c) and (d).}
    \label{fig:func_stat}
\end{figure}

We conduct a statistical analysis of the functional groups present in the dataset, as illustrated in Fig.~\ref{fig:func_stat}. For clarity, the top 25 functional group categories with the highest occurrence frequencies are counted and shown in (a) and (b). For the remaining categories, the top 35 functional groups with relatively higher frequencies are further analyzed and presented in (c) and (d). Our observations are as follows: (1) the distribution of functional groups in the training set and the test set is essentially consistent, which validates the rationality of the data partitioning; (2) the wide variety of functional group types demonstrates the diversity of the dataset; (3) the distribution of functional groups exhibits a long-tail characteristic.

\begin{figure}[t]
    \centering
    \includegraphics[width=0.99\linewidth]{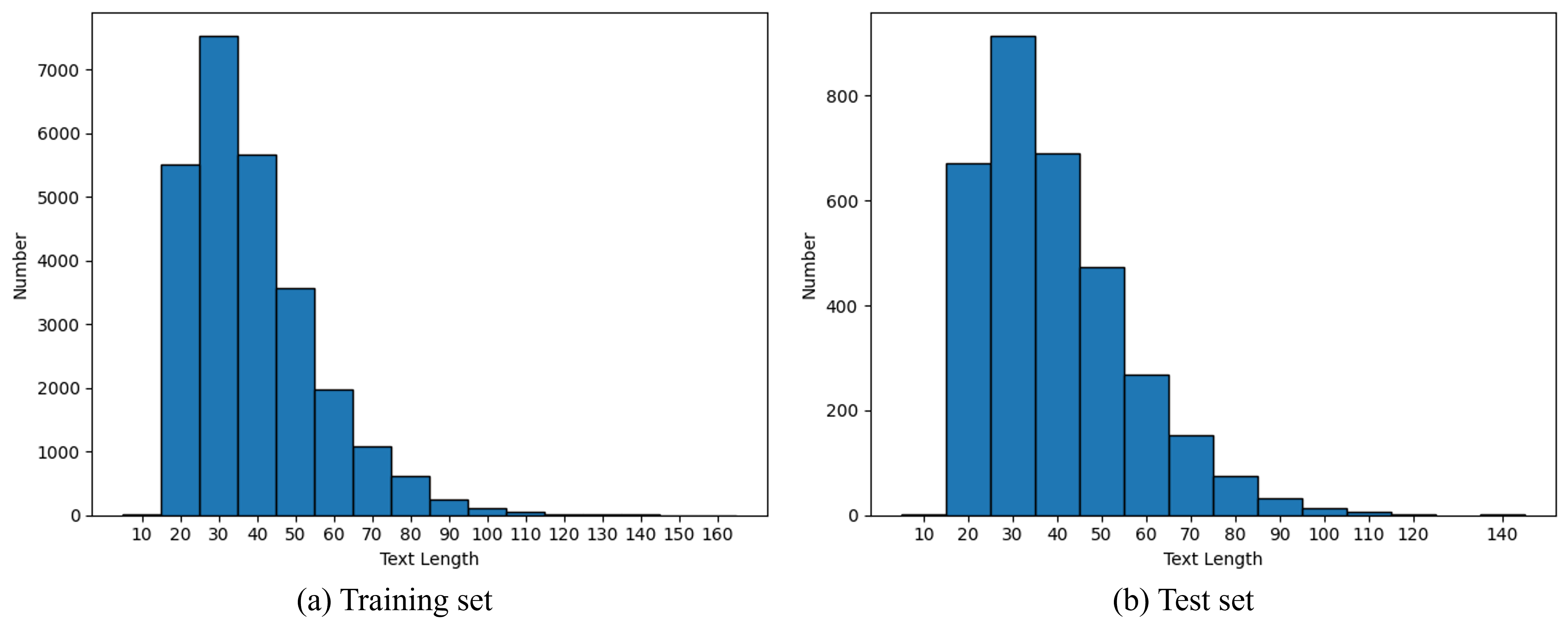} 
    \caption{\textbf{Statistical analysis on text length of molecule description.} (a) Statistical analysis of training set. (b) Statistical analysis of test set.}
    \label{fig:text_stat}
\end{figure}

We then perform a statistical analysis of the text length in the molecule description task, as shown in Fig.~\ref{fig:text_stat}. The results reveal that: (1) the majority of descriptions have word counts ranging from 20 to 30, with the vast majority containing fewer than 100 words; and (2) the text length distribution in the training set and the test set is essentially consistent.

\begin{figure}[t]
    \centering
    \includegraphics[width=0.99\linewidth]{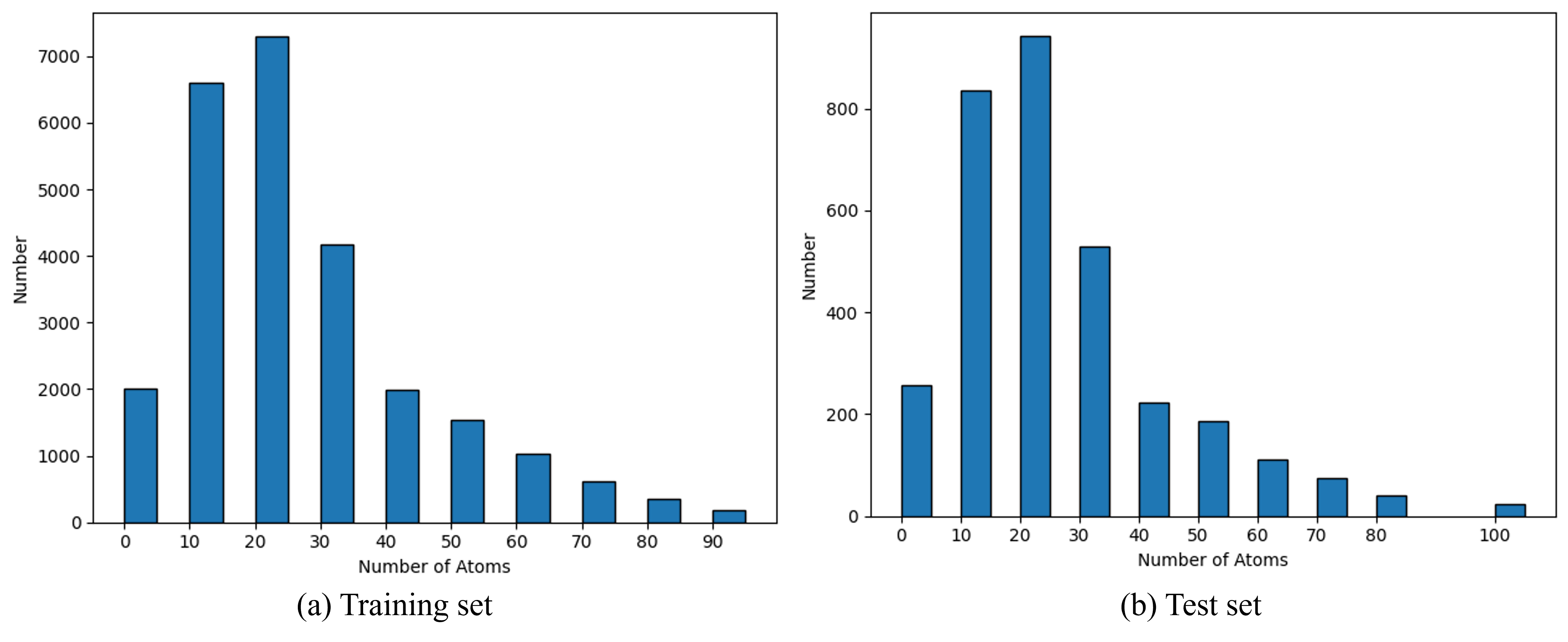} 
    \caption{\textbf{Statistical analysis on atom number of molecule.} (a) Statistical analysis of training set. (b) Statistical analysis of test set.}
    \label{fig:atom_stat}
\end{figure}

Finally, we assess the complexity of molecules by counting the number of atoms they contain, and the top 10 most frequent ranges are presented in Fig.~\ref{fig:atom_stat}. We observe that (1) the majority of molecules contain 20 to 30 atoms, with the vast majority having fewer than 100 atoms; and (2) the distribution of molecular complexity in the training set and the test set is essentially consistent.

\section{Performance Evaluation of DoubleCheck}
\label{sec:appendix_ocsr}

\begin{table*}[h]
\centering
\caption{\textbf{Comparison with others on patent and journal OCSR benchmarks.}}
\begin{tabular}{c|ccc|cc|ccc}
\hline
\multirow{2}{*}{Methods} & \multicolumn{3}{c|}{USPTO} & \multicolumn{2}{c|}{Staker\_p} & \multicolumn{3}{c}{ACS} \\ \cline{2-9} 
 & $Acc_s$ & $Acc_g$ & $Acc_c$ & $Acc_s$ & $Acc_g$ & $Acc_s$ & $Acc_g$ & $Acc_c$ \\ \hline
MolVec \cite{molvec} & 88.40 & 91.40 & - & 5.00 & 5.30 & 47.40 & 49.90 & - \\
OSRA \cite{osra} & 87.40 & 89.1 & - & 4.60 & 5.10 & 55.30 & 58.60 & - \\
Img2Mol \cite{img2mol} & 26.30 & 30.00 & - & 51.70 & 51.70 & 21.10 & 24.50 & - \\
DECIMER \cite{decimer} & 41.10 & 44.60 & - & 47.90 & 80.40 & 46.50 & 48.00 & - \\
MolScribe \cite{molscribe} & 92.57 & 94.53 & 82.79 & 65.04 & 87.87 & 71.30 & 75.53 & 39.07 \\ \hline
\textbf{DoubleCheck (Ours)} & \textbf{92.85} & \textbf{94.82} & \textbf{83.05} & \textbf{66.40} & \textbf{89.57} & \textbf{71.60} & \textbf{75.53} & \textbf{42.19} \\ \hline
\end{tabular}
\label{tab:ocsr_comparison}
\end{table*}

\textbf{Datasets.} We further compare DoubleCheck with existing state-of-the-art approaches on several public available benchmarks from previous work, including USPTO, ACS, and Staker\_p. \textit{USPTO} \cite{uspto} is a widely adopted patent benchmark comprising 5,719 molecular images. \textit{ACS} \cite{molscribe} is the only OCSR benchmark with 331 molecular images collected from American Chemical Society publications, which exhibit greater diversity in drawing styles. Following \cite{img2mol}, we also evaluate robust performance on the perturbed patent dataset \textit{Staker\_p} (29,228 images), which includes slight image rotations and shearing.

\textbf{Baselines.} We compare Double-Check with both rule-based and data-driven methods, including rule-based MolVec \cite{molvec} and OSRA \cite{osra}, and data-driven Img2Mol \cite{img2mol}, DECIMER \cite{decimer}, and MolScribe \cite{molscribe}. The performance comparison is reported in Tab.~\ref{tab:ocsr_comparison}.

\textbf{Evaluation metrics.} We evaluate the OCSR performance with two main metrics, exact matching accuracy $Acc_s$ and accuracy without considering chirality $Acc_g$. For unperturbed benchmarks, we further evaluate the recognition accuracy on chiral molecules $Acc_c$.

\textbf{Results and analysis.} Overall, our DoubleCheck outperforms existing methods on all these benchmarks. Benefit from the second look on ambiguous atoms, Double-Check achieves an absolute gain of $3.12\%$ on chiral molecules of ACS publications and exhibits stronger robust performance on Staker\_p ($1.36\% Acc_s$ and $1.7\% Acc_g$ over Molscribe). The SOTA performances (w./w.o. perturbation) demonstrate the effectiveness and advantage of the proposed attentive feature enhancement mechanism.

\section{Additional Case Studies}

\begin{figure}[h]
    \centering
    \includegraphics[width=0.99\linewidth]{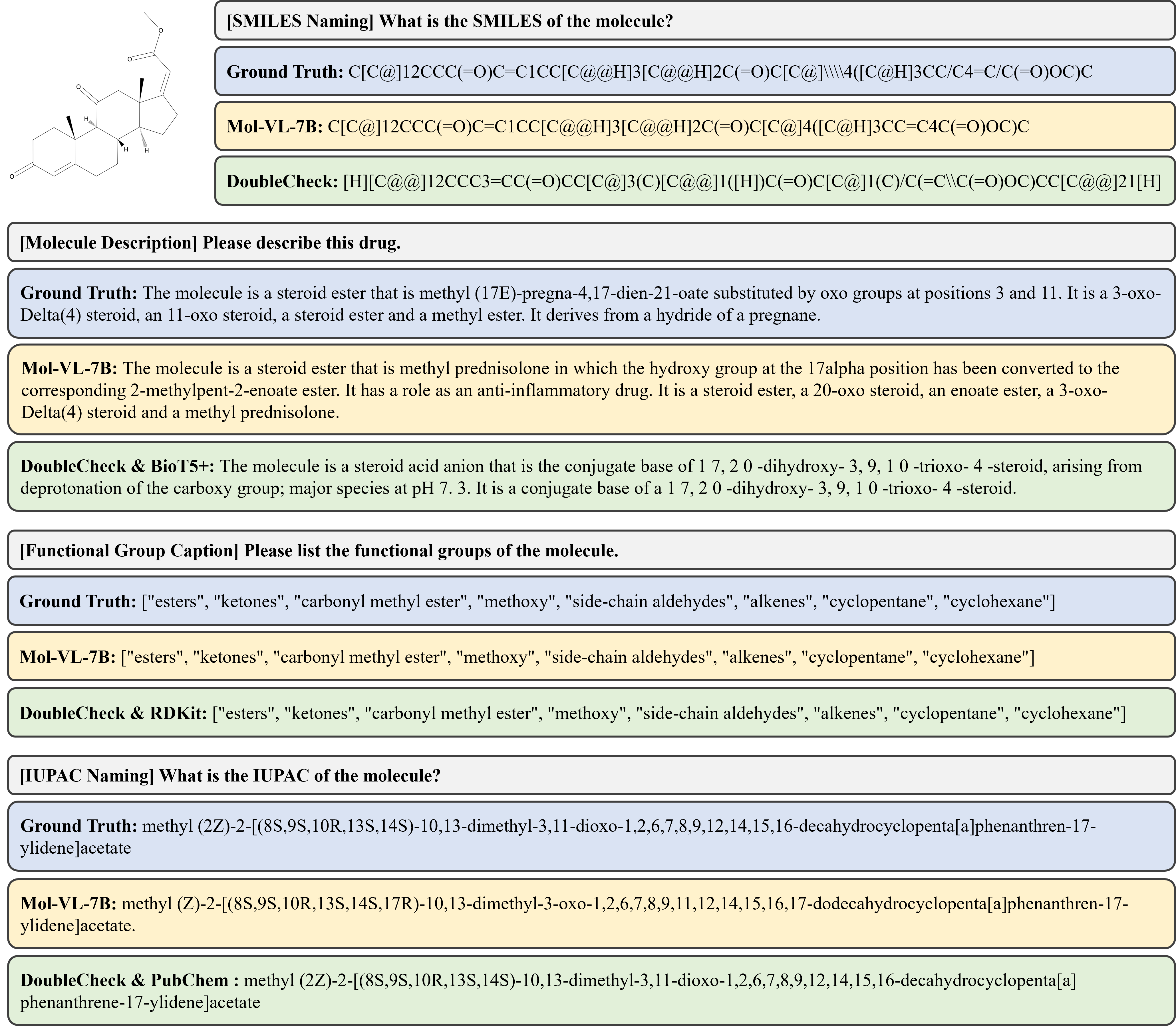} 
    \caption{\textbf{More case studies of optical chemical structure understanding on Vis-CheBI20.}}
    \label{fig:appendix_case}
\end{figure}

We present more case studies of optical chemical structure understanding in Fig.~\ref{fig:appendix_case}. We report the outputs of Mol-VL-7B and DoubleCheck-based approach.

\section{Error Analysis}
\label{sec:error}

\begin{figure}[h]
    \centering
    \includegraphics[width=0.99\linewidth]{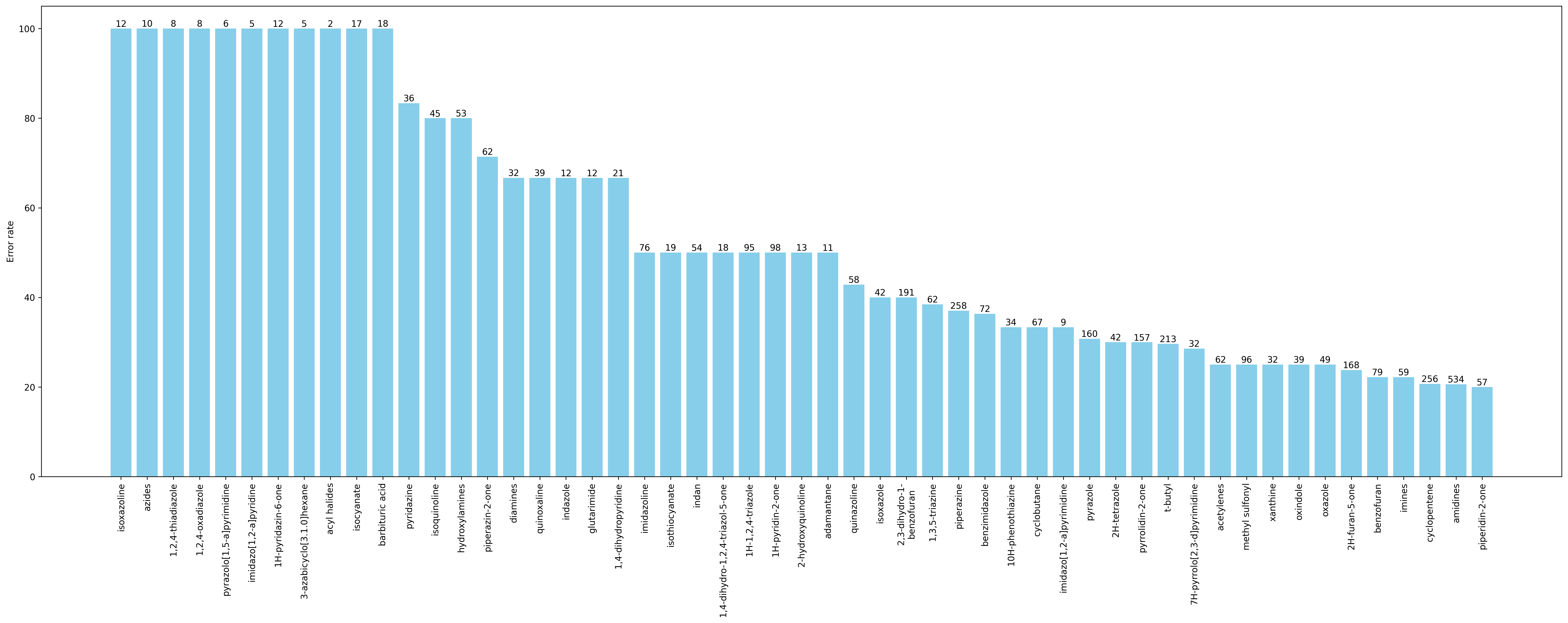} 
    \caption{\textbf{Error analysis of Mol-VL-7B on functional group caption.}}
    \label{fig:appendix_error}
\end{figure}

Mol-VL-7B achieves an F1 score of $97.32\%$ on the functional group caption task. To gain deeper insights, we conduct a detailed statistical analysis of the error cases, as presented in Fig.~\ref{fig:appendix_error}. We calculate the error rates of functional groups in the test set. For clarity, we only present the functional groups with error rates of $20\%$ or higher. Additionally, we label the frequency of these functional groups in the training set on the corresponding bars. It can be observed that functional groups with higher error rates generally appear less frequently in the training set. The long-tail distribution of the data impacts the model’s ability to accurately model functional groups. Therefore, we infer that scaling the training dataset will bring further improvement. How to achieve targeted data augmentation is also worth further investigation.

\section{Limitations and Broader Impacts}
\label{sec:appendix_limitation}

The proposed OCSU technologies, i.e., DoubleCheck and Mol-VL, have been successfully deployed in practical applications for patent analysis in drug discovery scenario, where it has significantly helped daily works. Nevertheless, the long-tail nature of the data presents challenges, particularly in comprehending certain uncommon chemical structures, which indicates potential for further performance enhancement. To address this, our future work will focus on enriching the dataset to better tackle corner cases. Additionally, given the recent advancements in the study of general VLM reasoning capabilities, we plan to incorporate optical chemical structure reasoning into our future research endeavors.

It is important to note that while OCSU technologies are primarily proposed for understanding tasks, subsequent molecule-centric scientific discoveries may potentially generate harmful molecules. We declare that OCSU technologies should be restricted to research purposes, and any further applications in molecule discovery should undergo comprehensive experiments.

\end{document}